\documentclass[letterpaper, 10 pt, conference]{ieeeconf}
\IEEEoverridecommandlockouts                              % This command is only needed if 
\usepackage{amsmath,amsfonts}
\let\labelindent\relax
\usepackage[T1]{fontenc}
\usepackage{algpseudocode}
\usepackage{algorithm}
\usepackage{array}
\usepackage{booktabs}
\usepackage{textcomp}
\usepackage{stfloats}
\usepackage{multirow}

\usepackage{url}
\usepackage{color}
\usepackage{xcolor}
\usepackage{wrapfig}

\usepackage{verbatim}
\usepackage{graphicx}
\usepackage{cite}

\usepackage{enumitem}

\DeclareMathOperator*{\argmin}{arg\,min}

\newcommand{\fref}[1]{Fig. \ref{#1}}
\newcommand{\sref}[1]{Section \ref{#1}}
\newcommand{\tref}[1]{Table \ref{#1}}

\usepackage{hyperref}
\hypersetup{
    colorlinks=true,
    bookmarksopen=true,
    bookmarksnumbered=true,
    citecolor=blue,
    urlcolor=black,
    pdfborder={0 0 0}
}

\title{
    \LARGE \bf
    iKap: Kinematics-aware Planning with Imperative Learning
}

\author{
    Qihang Li$^{1}$,
    Zhuoqun Chen$^{1}$,
    Haoze Zheng$^{1}$,
    Haonan He$^{2}$,
    Zitong Zhan$^{1}$,\\
    Shaoshu Su$^{1}$,
    Junyi Geng$^{3}$,
    Chen Wang$^{1}$
\thanks{*Corresponding Email: {\tt\small \{qihangl, chenw\}@sairlab.org}}
\thanks{$^{1}$Spatial AI \& Robotics (SAIR) Lab, University at Buffalo, USA.}
\thanks{$^{2}$Carnegie Mellon University, USA}
\thanks{$^{3}$Pennsylvania State University, USA}
}%

\begin{document}

\maketitle
\thispagestyle{empty}
\pagestyle{empty}

%%%%%%%%%%%%%%%%%%%%%%%%%%%%%%%%%%%%%%%%%%%%%%%%%%%%%%%%%%%%%%%%%%%%%%%%%%%%%%%%
\begin{abstract} 
Trajectory planning in robotics aims to generate collision-free pose sequences that can be reliably executed. Recently, vision-to-planning systems have gained increasing attention for their efficiency and ability to interpret and adapt to surrounding environments.
However, traditional modular systems suffer from increased latency and error propagation, while purely data-driven approaches often overlook the robot's kinematic constraints. This oversight leads to discrepancies between planned trajectories and those that are executable.
To address these challenges, we propose iKap, a novel vision-to-planning system that integrates the robot's kinematic model directly into the learning pipeline. iKap employs a self-supervised learning approach and incorporates the state transition model within a differentiable bi-level optimization framework. This integration ensures the network learns collision-free waypoints while satisfying kinematic constraints, enabling gradient back-propagation for end-to-end training.
% To address these limitations, we propose iKap, a novel system that integrates the robot's kinematic model directly into the learning pipeline. iKap employs a self-supervised learning approach and incorporates the state transition model into a differentiable optimization framework. This ensures that the network learns collision-free waypoints while maintaining compliance with kinematic constraints, enabling gradient backpropagation for end-to-end training. 
Our experimental results demonstrate that iKap achieves higher success rates and reduced latency compared to the state-of-the-art methods. Besides the complete system, iKap offers a visual-to-planning network that seamlessly works with various controllers, providing a robust solution for robots navigating complex environments.

% Our experimental results demonstrated that iKap achieves higher success rates and reduced latency. As a plug-and-play module, iKap can be easily integrated into various controllers, offering a robust solution for robots operating in complex and dynamic environments.
\end{abstract}

\section{Introduction} \label{sec:intro}
Path planning is a fundamental task in robotics, involving the determination of a collision-free trajectory that connects the robot's current position to the goal \cite{thrun2005probabilistic}. In dynamic and complex environments, robots must accurately perceive their surroundings and correlate environmental data with the planned path to ensure successful task execution \cite{SLAM_review}. Recently, with the development of perceptual systems, direct vision-to-planning systems have garnered increasing attention due to their potential efficiency and ability to interpret and adapt to surrounding environments \cite{sridhar_nomad_2023}.

Traditional vision-to-planning systems often employ a modular design where the planning module operates with an independent perception module that processes sensor data to generate environmental representations \cite{yang2022farplanner}. While this modular approach offers clarity and flexibility, it introduces several challenges. Communicating intermediate results between modules leads to potential latency, increased communication overhead, and higher storage demands \cite{Antonio_21_wind}. Moreover, errors in one module can propagate and amplify throughout the system, compromising overall performance \cite{SLAM_review}. 
% Traditional methods also often fail to extract rich semantic information from perception data, limiting the system's ability to understand and interpret complex environments .
\begin{figure}[!t]
    \centering
    \includegraphics[width=\columnwidth]{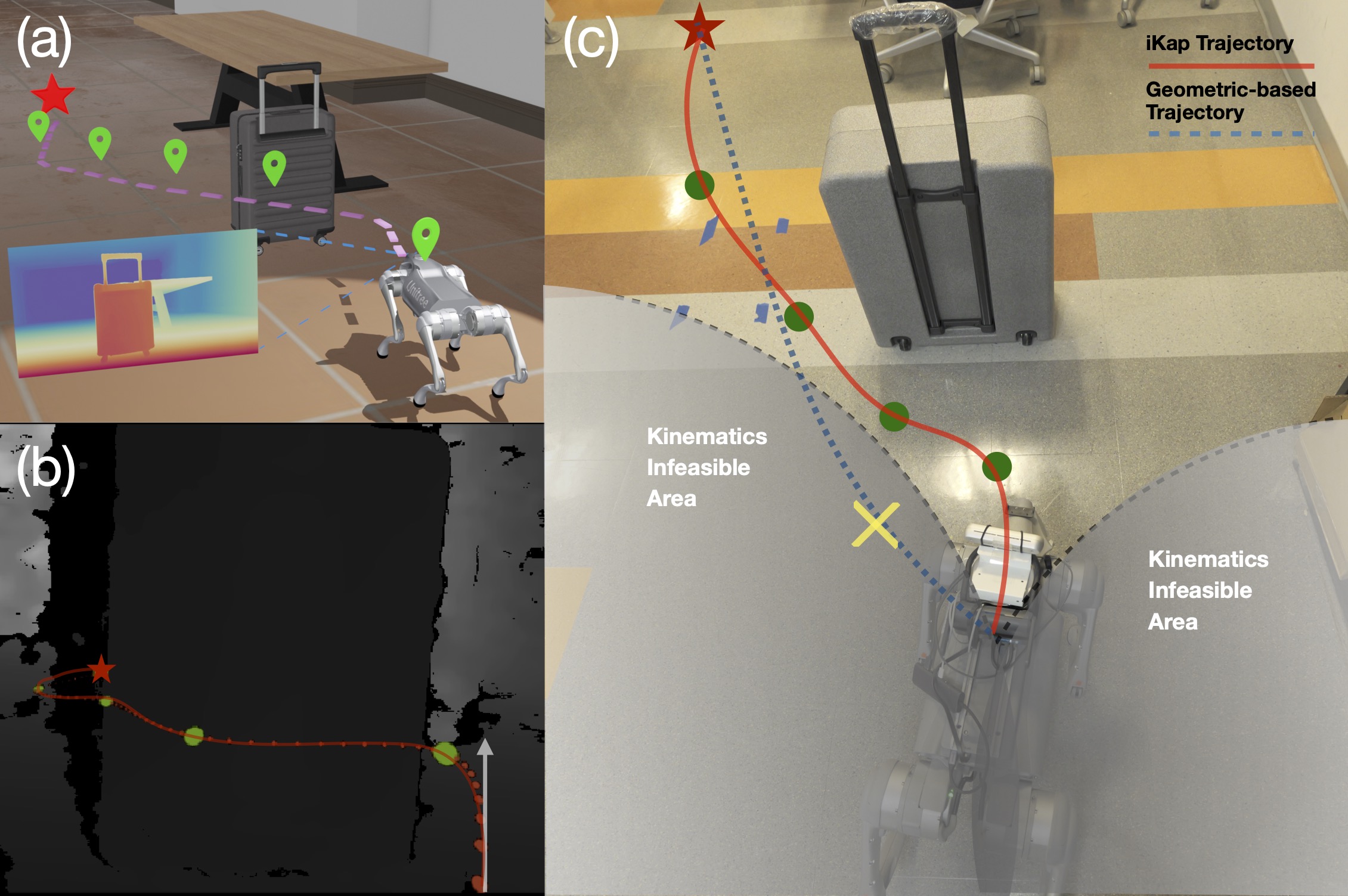}\\
    \caption{The kinematics-aware planner, iKap, incorporates kinematic state transition functions directly into the learning pipeline. This integration allows the network to learn feasible waypoint distributions and produce more executable trajectories. (a) The robot, constrained by a minimum turning radius, must reach the red star while avoiding obstacles. (b) A real-world depth image alongside the predicted trajectory. (c) A geometry-based planner may generate the blue trajectory, which does not satisfy the kinematic constraints, making it difficult to execute.}
    % \caption{The kinematics-aware planner iKap. Traditional planners often focus on geometry, neglecting the robot's kinematic constraints, making it difficult to execute trajectories under limitations like the minimum turning radius for vehicles. iKap embeds the kinematic state transition function equations into the learning pipeline, enabling the network to learn feasible waypoint distributions and generate more executable trajectories.}
    \label{fig:cover}
\end{figure}
Data-driven end-to-end approaches, particularly those utilizing deep learning, have obtained significant advancement in recent years \cite{sridhar_nomad_2023}. 
% Learning-based perception networks can leverage extensive training data to predict occluded regions and infer rich semantic information, enhancing accuracy and robustness in complex environments \cite{LeCun2015,Grigorescu2020}. 
Such methods directly map sensor inputs to control outputs,  eliminating the error propagation and latency inherent in modular systems \cite{Bojarski2016, amini_variational_2019, liu_efficient_2021, sridhar_nomad_2023}. 
However, purely data-driven methods often act as black boxes, require vast amounts of data, have low sample efficiency, and exhibit a significant sim-to-real gap. 
In addition, these approaches lack interpretability, making it challenging to incorporate explicit physical constraints or prior knowledge~\cite{Sauer2018}. Although recent works such as iPlanner~\cite{yang_iplanner_2023} have explored combining data-driven methods with trajectory optimization, they neglect the robot’s kinematics. This absence of physical constraints can result in generated trajectories that are difficult for low-level controllers to execute, leading to issues such as deadlocks, collisions, or suboptimal performance~\cite{Kuwata2009}. As illustrated in Figure~\ref{fig:cover}, for a robot with kinematic constraints like a minimum turning radius, the planned trajectory (blue curve) may not be executable by the robot.
% However, these approaches often act as black boxes, lacking interpretability and making it difficult to incorporate explicit physical constraints or prior knowledge about the robot's dynamics \cite{Sauer2018}. Moreover, purely data-driven methods such as iPlanner \cite{yang_iplanner_2023} that neglect the robots' kinematic model, may struggle to generalize to scenarios not represented in the training data, potentially leading to unsafe behaviors in untested environments.
% Moreover, in practice, the controller may struggle to accurately execute the planned trajectory, leading to issues such as deadlocks, collisions, or suboptimal performance during execution \cite{Kuwata2009}.

To address the issue of end-to-end systems generating infeasible paths, we argue that kinematic constraints should be incorporated into vision-to-planning networks while retaining the end-to-end learning architecture to avoid execution errors. 
To this end, we propose an Imperative learning-based Kinematics-aware Planning (iKap) system that integrates the robot’s kinematic model directly into the learning pipeline. This system supervises the network to learn kinematic-feasible trajectories through self-supervised bi-level optimization (BLO). The learning module of iKap is highly flexible and can be easily integrated with various controllers to plan executable trajectories. Our main contributions include:
% The absence of a kinematic model can lead to significant discrepancies between the planned and executable trajectories when these planning modules are integrated with control modules. 
% To this end, we propose an Imperative learning-based Kinematics-aware Planning (iKap) system to infuse the robot’s kinematic model into the learning pipeline. This system supervises the network to learn kinematically feasible trajectories through a self-supervised bi-level optimization (BLO).  The learning module of iKap is quite flexible, could be easily integrated with various controllers to plan executable trajectories. Our main contributions include:

\begin{itemize}
\item We integrate the robot's kinematic model into the training pipeline through a differentiable model predictive control (MPC) using a self-supervised BLO framework via imperative training. This enables the network to learn the kinematically feasible trajectories.
% \item We develop a learning pipeline that achieves gradient backpropagation from upper-level path planning to lower-level kinematics-model, facilitating end-to-end kinematics-aware vision-to-planning model.
\item We develop a learning pipeline that enables gradient backpropagation from the lower-level trajectory optimization to the upper-level network parameters, facilitating a kinematics-aware vision-to-planning model.
\item We present both simulation and real-world experiments to evaluate the proposed system in various environments. Results show that iKap achieves a higher success rate and lower tracking error than the baseline approach. 
\end{itemize}

\section{Related Work} \label{sec:related_work}

% iKap is related to data-driven planning and built upon imperative learning, which will be reviewed respectively.

\subsection{Data-driven Planner}

Our research builds upon iPlanner \cite{yang_iplanner_2023}, an end-to-end local planner that formulates trajectory planning as a BLO problem. iPlanner guides neural networks to plan collision-free paths trained with a mixed loss with imperative learning. However, it assumes that robots can perfectly execute trajectories optimized purely from geometric loss, neglecting kinematic feasibility and thus limiting real-world performance. Additionally, iPlanner simplifies the BLO to a single-level problem via a closed-form solution for the lower-level problem, enhancing computational efficiency but reducing generalizability to other constraints. ViPlanner \cite{roth_viplanner_2023} extends iPlanner by incorporating semantic information to construct cost map for local planning, yet it does not address the limitations regarding kinematic feasibility and generalizability.

Notable works like NoMaD \cite{sridhar_nomad_2023} utilize transformer-based architectures for action diffusion and motion prediction, integrating past observations with optional goal states to enhance decision-making. Similarly, ViNT \cite{vint} employs a transformer-based architecture to directly convert visual inputs into navigation decisions, enabling seamless adaptation across various environments and tasks. However, these methods lack physical representation during their training processes, potentially limiting their ability to account for dynamic and kinematic constraints.
Efforts to integrate physical models into data-driven planning include DIPP \cite{DIPP}, which combines a transformer with a differentiable trajectory model to plan based on road conditions and historical agent data. This approach requires supervised learning with labeled data, and the transformer architecture introduces latency, affecting real-time performance. PhysORD \cite{zhao_physord_2024} applies Lagrangian mechanics as a constraint for off-road driving motion prediction but does not incorporate visual data into the pipeline. Gao et al. \cite{Gao_techreport} demonstrated that embedding neural networks with safe corridors and trajectory optimization can efficiently generate collision-free and dynamically feasible trajectories, particularly for drones; however, it relies on supervised learning with results limited to simulations.

% Differentiable simulation has recently gained increasing attention for its ability to accelerate model learning by leveraging autodiff tools to compute the gradient propagation of system dynamics. Previous studies have demonstrated that integrating differentiable simulation into reinforcement learning systems significantly improves sample efficiency, enabling applications in vision-based control tasks \cite{visualRL}. Furthermore, recent work has integrated a simple point-mass model with a deep learning pipeline and achieved visual-based navigation tasks by introducing temporal gradient decay to mitigate gradient explosion \cite{newtonslaws}.

\subsection{Differentiable Optimization and Imperative Learning}
% Traditional robotics algorithms often rely on optimization techniques, which implicitly establish relationships between input and output data. Integrating these optimization-based algorithms into end-to-end systems can narrow the search space for neural networks, reduce the number of parameters, and improve computational efficiency \cite{Gao_techreport}.

% In this context, the differentiable optimization solver is critical, ensuring correct gradient propagation during the learning process. Amos et al. proposed implicit differentiation to manage gradient back-propagation for $\textit{argmin}$ problems, and integrating this process as a layer in neural networks. In their follow-up work, they introduced differentiable MPC, which plays a significant role in bridging MPC with deep learning frameworks \cite{optnet, amos2018differentiable}.

Traditional robotics algorithms often use optimization techniques to connect input and output data. Combining these optimization methods with end-to-end systems can reduce the search space for neural networks, lower the number of parameters, and boost computing efficiency \cite{Gao_techreport}. Here, a differentiable optimization solver is key to ensuring proper gradient flow during learning. Amos et al. suggested implicit differentiation to handle gradient back-propagation for $\textit{argmin}$ problems, embedding it as a layer in neural networks. Later, they introduced differentiable MPC, which helps link MPC with deep learning frameworks \cite{optnet, amos2018differentiable}.

Beyond efficiency, incorporating differentiable optimization into end-to-end training frameworks also preserves the interpretability of classical methods. Building on this idea, imperative learning has been proposed, embedding optimization-based algorithms as constraints and designing the upper-level loss based on the robot's state for self-supervised training \cite{wang_imperative_2024}. As a result, it formulates the network training problem as a BLO.
However, a key challenge lies in gradient backpropagation from the upper-level to the lower-level optimization, which many prior works have bypassed using closed-form solutions \cite{yang_iplanner_2023}, gradient assumptions at convergence \cite{fu_islam_2024, zhan2024imatching}, or discrete approximations \cite{guo2024imtsp, chen_ia_2024}. In our case, accurately computing gradients is critical due to the necessity of enforcing physical constraints in the BLO and the strong coupling between the two optimization levels. To address this, we employ a differentiable MPC \cite{pyposev06, ilqr}, demonstrating that imperative learning can effectively operate in constrained optimization scenarios.

\section{Method} \label{sec:method}
\subsection{Problem Formulation and System Overview} 

\begin{figure*}[t]
    \centering
    \includegraphics[width=\textwidth]{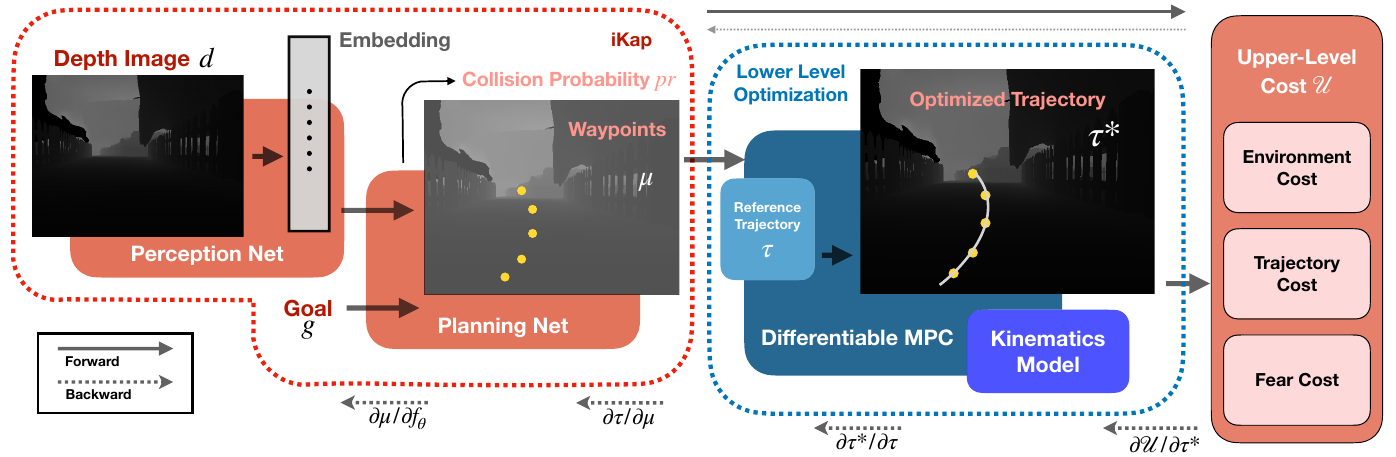}
    \caption{The training pipeline for the planner is based on BLO. In this framework, the networks, the optimized object, generate waypoints based on the given depth perception and goals. Then, the low-level trajectory optimization module tracks a kinematics-feasible trajectory. The embedded kinematics and gradients from the lower-level module are then utilized to supervise and train the networks.}
    \label{fig:flowchart}
\end{figure*}

At time stamp $t$, given observed depth image $\mathcal{D}_t$ and a target location $\mathbf{p}_t^g$ in the workspace $\mathcal{Q} \subset \mathbb{R}^2$, the planning problem is to find a trajectory $\mathbf{p}_{1:T+1}$ guiding the robot from the current position to the target while avoiding obstacles.

As illustrated in \fref{fig:flowchart}, our iKap consists of three key parts.
First, a ResNet\cite{resnet} front-end takes the depth observation $\mathcal{D}_t$ and encodes it into observation embedding. This embedding, combined with the target $\mathbf{p}_t^g$, is fed into the second planning network to predict a sparse key point trajectory $\boldsymbol{\mu}_{1:k}$ and collision probability $pr_t$.
Third, a differentiable MPC tracks a reference $\boldsymbol{\tau}:= \mathbf{x}_{1:T+1} \in \mathcal{X}^{T+1}$ interpolated from $\boldsymbol{\mu}_{1:k}$ with time horizon $T$ and outputs the optimized trajectory $\boldsymbol{\tau}^*_{1:T+1}$ and action $\mathbf{u}^*_{1:T}$ while ensuring the kinematic feasibility $\mathbf{x}_{t+1}=\mathbf{F}(\mathbf{x}_{t}, \mathbf{u}_{t})$. Here, $\mathbf{F}$ is the robot kinematics, $\mathbf{x}_{t}$ and $\mathbf{u}_{t}$ are the robot's states and control input in configuration space $\mathcal{X}$ and action space $\mathcal{A}$, respectively.
Finally, a well-designed traversability cost $\mathcal{U}$ is evaluated and backpropagated through both the differentiable MPC and the perception network to update the network parameters $\theta$. 
This process results in a BLO problem~\eqref{eq: bilevel}, where the perception and planning network is jointly optimized with the differentiable MPC. 
\begin{subequations}\label{eq: bilevel}
\begin{align}
    \min_{\theta} \quad &\mathcal{U}(\boldsymbol{\mu}_{1:k}, \boldsymbol{\tau}^{*}),\\
    \textrm{s.t.}\quad  & \boldsymbol{\tau}^{*} = \argmin_{\mathbf{u}_{1:T}} \mathcal{L}(\boldsymbol{\mu}_{1:k}, \mathbf{u}_{1:T}),\\
    &\textrm{s.t.}  \quad \mathbf{x}_{t+1} = \mathbf{F}(\mathbf{x}_{t}, \mathbf{u}_{t}),\\
    & \quad \quad \quad \ \mathbf{x}_0 = \mathbf{0},
\end{align}
\end{subequations}
where $\mathcal{L}$ denotes the cost of MPC. Intuitively, the upper-level cost \(\mathcal{U}\) is iteratively updated based on the lower-level MPC, which can be optimized without providing labels, enabling a self-supervised learning process. Moreover, during the optimization, \(\mathcal{U}\) consistently incorporates the robot's kinematic models through the lower-level MPC, ensuring a kinematics-aware planning network. We next present the details of the planning module and MPC in \sref{sec:planning} and \sref{sec:mpc}, respectively. The upper-level cost together with the method solving BLO will be illustrated in \sref{sec:cost}.

This bi-level design separates the neural network and the traditional optimization-based approach at key-point level, allowing each to focus on its strengths—the neural network processes unstructured image inputs to predict the key-point path, while MPC generates actions and incorporates interchangeable kinematics constraints.

\subsection{Perception and Planning Networks} \label{sec:planning}
The network consists of two components: the perception network, which processes depth images, and the planning network, which generates a trajectory based on the goal position and the output of the perception network.
% The network has two parts: the perception network, which handles depth images, and the planning network, which creates a trajectory using the goal position and the perception network's output.
\subsubsection{Perception Network}
We use a similar network structure as in \cite{yang_iplanner_2023} for perception.
At each timestep $t$, the robot's sensor captures depth images $\mathcal{D}_t$. These images are processed by the perception network, which employs an enhanced ResNet architecture \cite{resnet} to extract high-dimensional spatial features. After the encoder stage, the depth images are transformed into a higher-dimensional representation $\mathcal{O}_t \in \mathbb{R}^{C \times M}$, where $C=512$ denotes the number of channels and $M=240$ represents the dimension of the feature space. 

\subsubsection{Planning Network}
The planning module employs the embedding from perception network $\mathcal{O}_t \in \mathbb{R}^{C \times M}$ together with the goal position $\mathbf{p}_t^{g} \in \mathbb{R}^2$ to generate waypoints. To align the goal position with the embedding's dimensionality, $\mathbf{p}_t^{g}$ is mapped into a higher-dimensional space $\mathcal{P}_t \in \mathbb{R}^{C^* \times M}$ via a simple multilayer perceptron (MLP). The embedding $\mathcal{O}_t$ and the transformed goal position $\mathcal{P}_t$ are then concatenated to form the final input $\hat{\mathcal{O}} \in \mathbb{R}^{C' \times M}$, where $C' = C + C^*$. The planning network, consisting of CNN and MLP with activation functions, predicts waypoints $\mathcal{K}_{t} \in \mathbb{R}^{n \times 2}$, where $n$ represents the number of key points.

\subsection{Differentiable Model-Predictive Control} \label{sec:mpc}
We leverage a differentiable MPC module to incorporate the robot kinematic model into the training pipeline. The MPC can generate trajectories that comply with kinematic constraints, which can be formulated as:
\begin{align}\label{dmpc}
    \boldsymbol{\tau}^* = &\ \argmin_{\mathbf{x}_{1:T}, \mathbf{u}_{1:T}} \  \sum_{t=0}^{T-1} (\tilde{\mathbf{x}}^\top_t\mathbf{Q}_t\tilde{\mathbf{x}}_t + \mathbf{u}_t^\top \mathbf{R}_t \mathbf{u}_t) + \tilde{\mathbf{x}}^\top_T\mathbf{Q}_T\tilde{\mathbf{x}}_T \\
    &\quad \textrm{s.t.} \quad \mathbf{x}_{t+1} = \mathbf{F}(\mathbf{x}_t, \mathbf{u}_t),\\
    &\quad \quad \quad~ \mathbf{x}_0 = \mathbf{0},
\end{align}
where $\tilde{\mathbf{x}}_t = \mathbf{x}_t - \mathbf{x}_t^{\text{ref}}, t=0,\cdots,T$ is the error state between the current state $\mathbf{x}_t$ and reference trajectory $\mathbf{x}^{\text{ref}}$; $\tilde{\mathbf{x}}^\top_t\mathbf{Q}_t\tilde{\mathbf{x}}_t + \mathbf{u}_t^\top \mathbf{R}_t \mathbf{u}_t$ is the stage cost balancing the state and control effort; $\tilde{\mathbf{x}}^\top_T\mathbf{Q}_T\tilde{\mathbf{x}}_T$ is the terminal cost; $\mathbf{Q}_t$, $\mathbf{R}_t$,$\mathbf{Q}_T$ are the weight matrices.
Specifically, the reference trajectory is obtained by interpolating the waypoints predicted by the network. 
% The classic Dubins car \eqref{dubin_car} can be used to describe kinematic model of a legged robot.
% Limited by the differentiable MPC solver, we use the classic Dubins car model \eqref{dubin_car} to describe the kinematic model of a legged robot. More complex kinematic models can be integrated into the pipeline in a similar approach.
To show the capability of the differentiable MPC solver, we employ the classic Dubins car model \eqref{dubin_car} to represent the kinematic model of a legged robot as an example. Notice that our approach can integrate more complex kinematic models into the pipeline in a similar manner.
\begin{equation}\label{dubin_car}
\dot{x}_t =v \cos \theta_t, \quad
\dot{y}_t =v \sin \theta_t, \quad
\dot{\psi}_t = u_t,
\end{equation}
where $\mathbf{x}_t = [x_t, y_t, \psi_t]^\top$ is the robot states representing the position and heading and $\mathbf{u}_t = [v, u_t]^\top$ is the control input including the robot speed and turn rate. We then discretize~\eqref{dubin_car} and use it as the kinematic constraint for~\eqref{dmpc}.
To avoid motion singularity and gradient instability for complex trajectories, we use $\mathbf{SE}(2)$ Lie group operation for robot kinematics.

Since $\boldsymbol{\tau}^*$ is a function of the reference path $\boldsymbol{\tau}$, which is a function of the predicted key point path $\boldsymbol{\mu}_{1:k}$ or $f_{\theta}$, and will be passed to the upper level optimization for the final end-to-end training, the critical part is to calculate the gradient $\frac{\partial \boldsymbol{\tau}^*}{\partial \boldsymbol{\tau}}$. The traditional unrolling approach maintain the computational graph throughout the entire iteration process \cite{wang_imperative_2024}, which poses significant computation burden when dealing with complex problems. It may run into divergent or vanishing. Instead, we employ the implicit function differentiation theorem and leveraging the KKT condition of~\eqref{dmpc} at the optimal point to compute the gradients of the parameters~\cite{amos2018differentiable}. Thus, there is no need for explicit unrolling of the entire iteration process. 
In practice, we use the differentiable MPC module in our robot learning library PyPose \cite{wang2023pypose, pyposev06}, which solves the MPC by the iterative Linear Quadratic Regulator algorithm \cite{ilqr}. PyPose MPC adopts a more general and problem-agnostic automatic differentiation based solution that introduces just one additional optimization iteration at the optimal point in the forward pass. This allows automatic gradient calculation during the backward pass without the need to differentiate through the entire unrolled chain or compute problem-specific analytical derivatives.

\subsection{Upper-level Cost Design and Optimization} \label{sec:cost}

To achieve self-supervised learning with BLO, an important component is the upper-level cost design and accurate gradient back-propagation.
We formulate the upper-level cost function as a weighted summation of fear cost $\mathcal{C}^{\mathcal{F}}$, environment cost $\mathcal{C}^{\mathcal{E}}$, and trajectory cost $\mathcal{C}^{\mathcal{T}}$:
\begin{equation}
\mathcal{U}(\boldsymbol{\tau}, \boldsymbol{\mu}_{1:k})=\alpha \cdot \mathcal{C}^{\mathcal{F}}(\boldsymbol{\tau}) + \beta \cdot \mathcal{C}^{\mathcal{E}}(\boldsymbol{\tau}, \boldsymbol{\mu}_{1:k}) + \gamma \cdot \mathcal{C}^{\mathcal{T}}(\boldsymbol{\tau}).
\end{equation}
As described in \sref{sec:planning}, the planner network predicts a collision probability $pr_t$ along with the key point path. The fear cost $\mathcal{C}^{\mathcal{F}}$ in \eqref{eq:fear} computes the binary cross entropy (BCE) as the task-level cost to avoid collisions.
\begin{equation}\label{eq:fear}
\mathcal{C}^{\mathcal{F}} = \begin{cases}\operatorname{BCELoss}(pr_t, 0.0) & \mathbf{p} \cap \mathcal{Q}_{obs}  \neq \emptyset, \\ \operatorname{BCELoss}(pr_t, 1.0) & \text{ otherwise}.\end{cases}
\end{equation}
The cost $\mathcal{C}^{\mathcal{E}}$ in \eqref{eq:environment} is the sum of the shortest distance between the trajectory points to the environment. Each waypoint projected on a Euclidean Signed Distance Field (ESDF) map.
% represents the geometric cost between the environment obstacle and the trajectory during planning. The shortest distance between the robot and the environment, i.e., ESDF map, is used to ensure safety.
\begin{equation}\label{eq:environment}
\mathcal{C}^{\mathcal{E}} = \frac{1}{k} \sum_{i=1}^k \rm{ESDF}_\mathcal{E}(\boldsymbol{\mu}_i) + \frac{1}{T} \sum_{i=1}^T \rm{ESDF}_\mathcal{E}(\mathbf{p}_i).
\end{equation}
The trajectory cost $\mathcal{C}^{\mathcal{T}}$ in \eqref{eq:trajectory}, consisting of three terms, is to evaluate the trajectory shape and trackability.
\begin{equation}\label{eq:trajectory}
\begin{aligned}
    \mathcal{C}^{\mathcal{T}} 
    =& \gamma_{1}\cdot \log (\|\boldsymbol{\mu}_k-\mathbf{p}_t^g\|_2+1.0) \\
    +& \gamma_{2}\cdot \frac{1}{T-1}\sum_{i=2}^{T} \|\frac{i\cdot\boldsymbol{\mu}_k}{T}-\mathbf{p}_i\|_2 \\
    +& \gamma_{3}\cdot \frac{1}{T}\sum_{i=2}^{T+1} \|\tilde{\mathbf{x}}_{i}\|_2.
\end{aligned}
\end{equation}
The first term encourages the planner to generate a trajectory close to the goal, and the design of $\log(\text{L2-norm} + 1)$ ensures that the loss is always positive while providing a smooth gradient. The second term measures the deviation from a straight, uniformly distributed path, ensuring that the trajectory is evenly spaced. This design also helps smooth out disorganized paths, preventing the learning process from becoming trapped in local minima. The final term focuses on the relationship between the reference and kinematics-feasible trajectory. Minimizing it allows the network to learn the distribution of kinematically feasible trajectories.

To solve the BLO in \eqref{eq: bilevel}, the gradient of the upper-level cost must be propagated back through the lower-level system by applying the chain rule, and the network parameter is updated using gradient descent:
\begin{align}
\nabla_\theta \mathcal{U} &=(\frac{\partial \mathcal{U}}{\partial \boldsymbol{\mu}} +\frac{\partial \mathcal{U}}{\partial \boldsymbol{\tau}^*} \boldsymbol{\frac{\partial \tau^*}{\partial \boldsymbol{\tau}}} \boldsymbol{\frac{\partial \tau}{\partial \boldsymbol{\mu}}})\frac{\partial \boldsymbol{\mu}}{\partial \theta}\\
\theta_{t+1} &= \theta_{t} - \alpha\cdot\nabla_\theta \mathcal{U}.
\label{eq:gradient}
\end{align}
where $\alpha$ is the learning rate. Notice that most of the gradient terms in \eqref{eq:gradient} can be computed automatically in PyTorch. However, the relationship between the reference trajectory $\boldsymbol{\tau}$ and the optimized trajectory $\boldsymbol{\tau}^*$ is determined through an $\rm{argmin}$ optimization, which is difficult to compute. Using the differentiable MPC in PyPose \cite{wang2023pypose}, we can resolve this through implicit differentiation.
%by iteratively linearizing the state transition function along the trajectory and minimizing the loss. Through this method, we obtain an approximated implicit differentiation between $\boldsymbol{\tau}$ and $\boldsymbol{\tau}^*$.

\section{Experiments} \label{sec:experiments}
\subsection{Implementation, Platforms, and Baselines}

\begin{figure}[t]
    \centering
    \includegraphics[width=\linewidth]{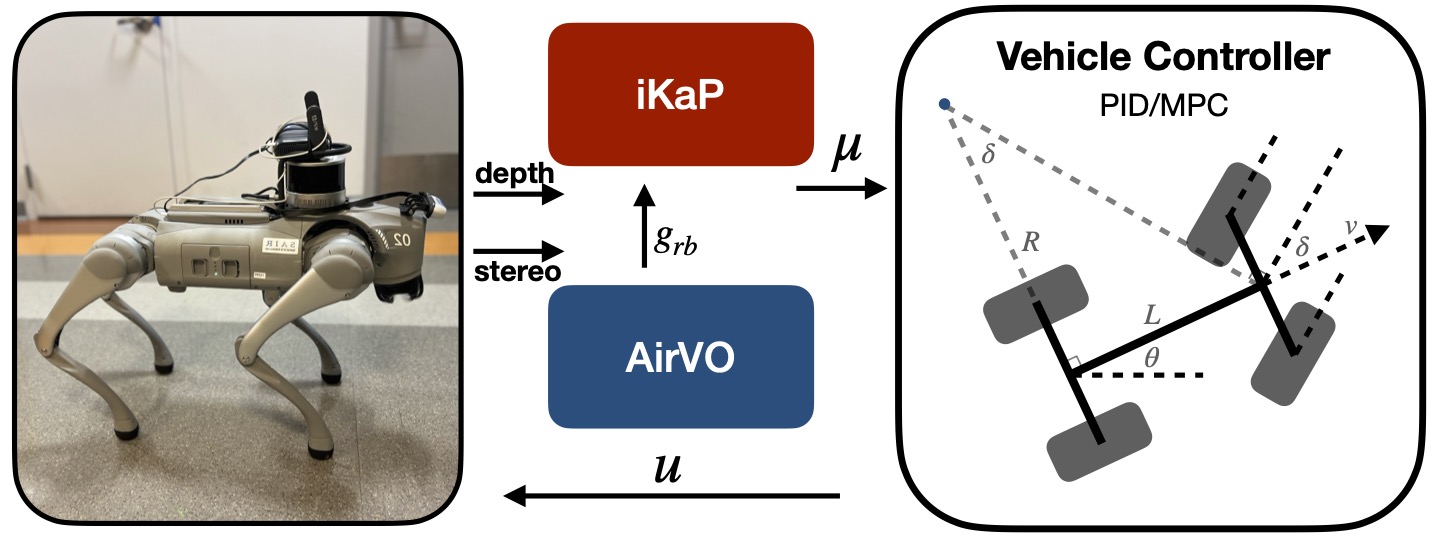}
    \caption{The real-world experiment pipeline. The experiments are conducted on the Unitree Go2 robot dog. We use AirVO to estimate the robot's odometry, command the goal waypoint to the robot, and input this information into iKap along with depth images captured by the RealSense D435i camera.}
    \label{fig:realrb_pipeline}
\end{figure}

We next evaluate the performance of iKap in both simulated and real-world settings. Specifically, we employ the simulated environments for autonomous exploration provided in \cite{cao2021autoEnv}, running on a desktop equipped with an Intel i7-12700k CPU and an Nvidia RTX 3090 GPU. For real-world experiments, we deploy the Unitree GO2 legged robot as shown in \fref{fig:realrb_pipeline}, outfitted with an Intel RealSense D435i camera that offers depth and stereo perception at 30 Hz. To assess generalization capability, we collect training data only from simulations, where the robot is joystick-controlled in four different environments: forests, garage, indoor, and Matterport3D \cite{m3d}. We implement the stereo odometry solution AirVO \cite{AirVO} to track the robot's position and generate an ESDF map. The training process, executed on an A6000 GPU, takes approximately six hours. 

As a baseline model, we adopt the pre-trained iPlanner \cite{yang_iplanner_2023}. To facilitate clarity, we summarize the following settings both for baselines and our proposed method:

% \item \textbf{Ours:} As is presented in Sec.~\ref{sec:method}, \SI{50}{\percent} \textit{during training}.

\begin{enumerate}[leftmargin=*]
    \item \textbf{iKap:} This is \textbf{Our} proposed planning network as is presented in \sref{sec:method}. It is jointly trained with Dubins car model and allows plug-and-play replacement by different low-level controllers designed for other kinematics.
    \item \textbf{iKap+PID:} This setting is at test time. We use a PID controller to track iKap planned reference trajectory.
    \item \textbf{iKap+MPC:} This setting is at test time. We use an MPC controller to track iKap planned reference trajectory.
    \item \textbf{iPlanner} This is the baseline setting and it differs from \textbf{Ours} in that it is trained without kinematics embeddings.

\end{enumerate}

\subsection{System Performance}

\begin{figure*}[t]
    \centering
    \includegraphics[width=\textwidth]{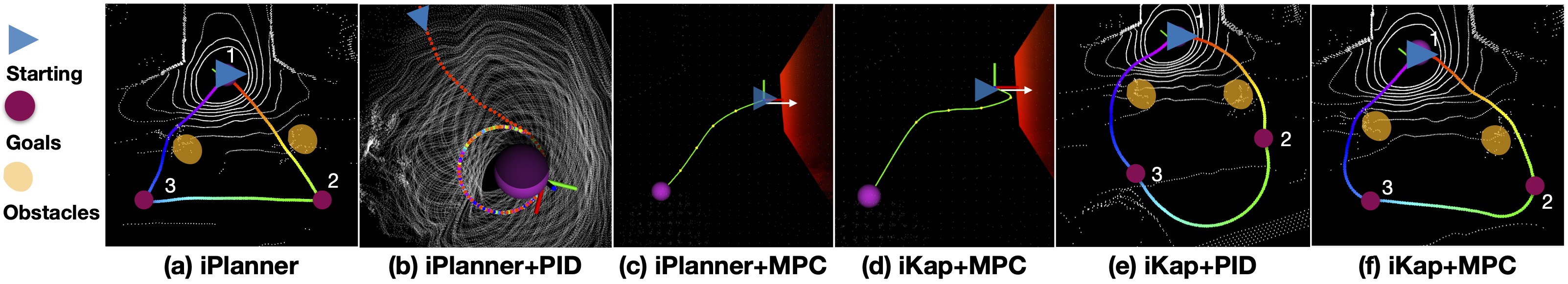}
    \caption{
    Qualitative performance.
    \textbf{(a) iPlanner without controller:} The planner generates sharp-angled trajectories that are difficult to execute.
    \textbf{(b) iPlanner with PID controller:} Without considering kinematics, iPlanner accumulates significant tracking errors, making control challenging. Near the goal, these errors cause iPlanner to get stuck in a looping behavior.
    \textbf{(c, d) U-turn tests:} iPlanner (c) generates backward trajectories, which are difficult for the MPC controller to execute. In contrast, iKap+MPC in (d) plans feasible trajectories that the controller can follow.
    \textbf{(e, f) Trajectory Tracking:} Using PID (e) and MPC (f) to track the waypoints generated by iKap. The robot can navigate robustly.
    }
\label{fig:sim}
\end{figure*}

\begin{table}[t]
\centering
\caption{Average Tracking Error (unit: m) ($\downarrow$)}
% (Minimum turn Radius = 1.46m)
\vspace{-10pt}
\resizebox{\linewidth}{!}{%
\begin{tabular}{c|l|cccc}
\toprule
% \multicolumn{6}{c}{\textbf{Average Tracking Error} (m) ($\downarrow$)} \\ 
\textbf{Controller} & \textbf{Method} & \textbf{Forests} & \textbf{Garage} & \textbf{Indoor} & \textbf{Matterport}\\
\midrule
\multirow{2}{*}{PID} 
 & iPlanner & 0.2084 & 0.2542 & 0.3157 & 0.3594\\
 & iKap     & \textbf{0.1946} & \textbf{0.2394} & \textbf{0.2302} & \textbf{0.3387}\\
 % \addlinespace
 \midrule
% \addlinespace
\multirow{2}{*}{MPC} 
 & iPlanner & 0.1202 & 0.2137 & 0.1218 & 0.3025 \\
 & iKap     & \textbf{0.0832} & \textbf{0.1446} & \textbf{0.1042} & \textbf{0.2568} \\
\bottomrule
\end{tabular}
}
\label{table:tracking_error}
\end{table}

% \begin{table}[H]
% \centering
% \caption{Average Tracking Error (Radius = 1.46m)}
% \resizebox{0.5\textwidth}{!}{%
% \begin{tabular}{l|l|cccc|c}
% \toprule
% \textbf{Controller} & \textbf{Method} & \textbf{Forests} & \textbf{Garage} & \textbf{Indoor} & \textbf{Matterport} & \textbf{Overall} \\
% \midrule
% \multirow{2}{*}{PID} 
%  & iPlanner & 0.2084 & 0.2542 & 0.3157 & 0.3594 & 0.2844 \\
%  & iKap     & \textbf{0.1946} & \textbf{0.2394} & \textbf{0.2302} & \textbf{0.3387} & \textbf{0.2507} \\
%  \addlinespace
%  \hline
% \addlinespace
% \multirow{2}{*}{MPC} 
%  & iPlanner & 0.1202 & 0.2137 & 0.1218 & 0.3025 & 0.1895 \\
%  & iKap     & \textbf{0.0832} & \textbf{0.1446} & \textbf{0.1042} & \textbf{0.2568} & \textbf{0.1472} \\
% \bottomrule
% \end{tabular}
% }
% \label{tab:merged_tracking_error}
% \end{table}

% \begin{table}[H]
% \centering
% \caption{PID Tracking Error}

% \begin{tabular}{lcccccc}
% \toprule
% \textbf{PID} & \multicolumn{5}{c}{\textbf{Average Tracking Error} (Radius = 1.46m)} \\
% \cmidrule(lr){2-6}
% & Forests & Garage & Indoor & Matterport & Overall \\
% \midrule
% \textbf{iPlanner} & 0.2084 & 0.2542 & 0.3157 & 0.3594 & 0.2844 \\
% \addlinespace[0.5em]
% \textbf{CiPlanner} & \textbf{0.1946} & \textbf{0.2394} & \textbf{0.2302} & \textbf{0.3387} & \textbf{0.2507} \\
% \bottomrule
% \end{tabular}

% % \addlinespace[0.5em]
% \label{tab:pid}
% \end{table}

To demonstrate the flexibility and generalization capability of the proposed method, we conduct zero-shot transfer tests by applying a pre-trained model, initially embedded with a Dubins car model, to an unseen low-level bicycle model both in simulation and real world. Additionally, we employ two different controllers, PID and MPC to follow the reference trajectories generated by the iKap module.

\subsubsection{Qualitative Result}

An important factor in evaluating the effectiveness of the reference trajectory generated by the planner, is that it can be successfully tracked by different low-level controllers and robots with varying kinematics. iPlanner's assumptions are highly limited, as it only considers geometric constraints of the trajectory. \fref{fig:sim}~(a) illustrates the trajectory produced by iPlanner when the target point is at the vertex of a star-shaped pattern. To follow such a trajectory, the robot must turn in place before executing the next segment, which is difficult for robots with other configurations. \fref{fig:sim}~(b) shows the compounding error issue when a robot modeled with bicycle kinematics uses a PID controller to track the iPlanner trajectory. If the tracking error falls within a critical range smaller than the robot's minimum turning radius, the robot may continuously circle around the target point, leading to tracking failure. In U-turn tests where the goal position is behind the robot, iPlanner generates a trajectory like the one in \fref{fig:sim}~(c), whereas iKap generates a trajectory with a turning angle, as shown in \fref{fig:sim}~(d), providing the MPC controller with a more kinematically feasible starting point, thus reducing the gap between the planner and controller. \fref{fig:sim}~(e) illustrates the actual trajectory executed using a PID controller when static obstacles are present along the path. Similarly, \fref{fig:sim}~(f) depicts the trajectory executed by iKap+MPC, demonstrating stable planning capabilities and effective obstacle avoidance.

% When integrating with the controller, iPlanner encountered issues due to discrepancies between the planned and executable trajectories. \fref{fig:sim} shows a qualitative result. \fref{fig:sim}~(a) depicts the original iPlanner trajectory, which only considers geometry and plans an untrackable path with large angle variations. Both PID and MPC controller are implemented to track the reference trajectory commanded by the upper-level module. In \fref{fig:sim}~(b), a PID controller is used to track the planned trajectory. The PID controller, which performs closed-loop control based on error, is designed for fast but lower-accuracy control. During execution, the errors arising from the gap between the planned and executable trajectories disrupt the controller, preventing the robot from accurately reaching the goal position. In some cases, the robot gets stuck due to limitations in its turning radius. \fref{fig:sim}~(c) and \ref{fig:sim}~(d) show the results of trajectory tracking using the MPC controller for iPlanner and iKap, respectively. The results indicate that when facing sharp turns or U-turns, iPlanner generates untrackable trajectories, causing the MPC to get stuck. In contrast, the trajectories planned by iKap remain executable in these situations. \fref{fig:sim}~(e) and \ref{fig:sim}.(f) show the results of iKap’s planner working in conjunction with the PID and MPC controllers, respectively. iKap demonstrate the ability to perform kinematically feasible planning and obstacle avoidance in complex environments, even under significant planning angle variations.
\begin{figure}[H]
    \centering
    \includegraphics[width=\linewidth]{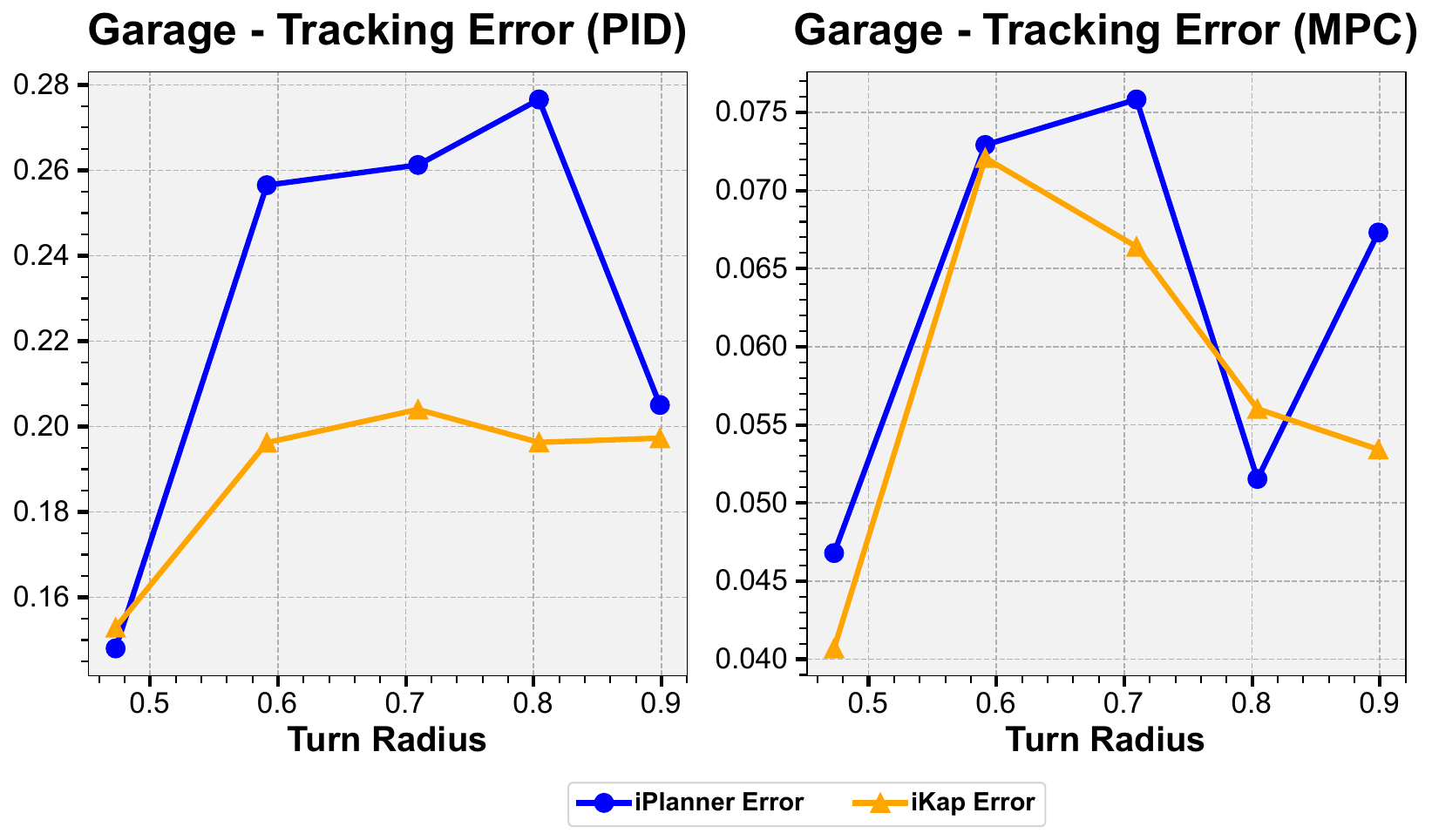}
    \caption{The trajectory tracking error against the minimum turning radius by PID and MPC controller in Garage simulation environment}
\label{fig:track}
\end{figure}

\subsubsection{Kinematic Feasibility}
To evaluate the kinematic feasibility of different planners, we next show the tracking errors of generated paths executed by the controllers. To this end, we collect 200 depth images and goal pairs from the four simulation environments.
We then calculate the mean squared error between the planned trajectories and trajectory executed by a controller, assuming a minimum turning radius of 1.48 meters.
As shown in \tref{table:tracking_error}, iKap reduces the tracking error by 11.85\% with the PID controller and by 22.34\% with the MPC controller, compared to iPlanner, across the four environments.
This demonstrates significant advantages of iKap compared to iPlanner when given kinematic constraints.

To further show the performance of kinematic feasibility, we illustrate the tracking error in terms of turning radius by adjusting the maximum steering angle of the front wheels. 
As shown in \fref{fig:track}, iKap can easily incorporate different minimum turning radius constraints and shows fewer tracking errors than iPlanner, demonstrating the flexibility of iKap in terms of kinematics-aware planning.
This experiment demonstrates that iKap is robust to parameter variations, consistently generating kinematics-aware waypoints that are easier to track under different constraint conditions. 

\subsubsection{Navigation Test}
% The previous experiment focused on analyzing the traversability of the predicted trajectory, but in dynamic tasks, planning becomes more complex. 
We next evaluate the planners' performance in a navigation task.
To this end, we calculate the failure rate of a planner, where a failed attempt is defined as any instance where a collision with the environment occurred, the robot became deadlocked during planning, or the trajectory could not be executed by the controller.
Specifically, we collected total 100 initial pose and goal pairs across four scenarios: "forest" (17) with dense trees for testing obstacle avoidance, "garage" (23) with partitioned warehouses, "indoor" (31) featuring narrow corridors, and "campus" (29) with diverse mixed terrains. We then used both iPlanner and iKap to generate local trajectories and navigate the robot to its destination in each environment. 
As shown in \tref{tab:success_rate}, iKap demonstrates superior performance in three environments.
It can be seen that failure rate of iKap is slightly higher than in the ``Garage'' environment. This is because the robot often faces obstructed vision due to the presence of numerous walls and blind spots, while iKap is sensitive to the kinematic infeasible region.

% % Although the kinematic constraints helped iKap narrow down the search space and plan kinematically feasible trajectories more effectively, they also made iKap more prone to overfitting to infeasible trajectories in situations with insufficient information. 
% Additionally, when iPlanner was integrated into the control pipeline with kinematic constraints, planning difficulties also arose due to the restricted feasible region.
\begin{table}[t]
\centering
% \caption{Success Rate Comparison}
\caption{Success Rate (\%) (with MPC Controller) ($\uparrow$)}
\vspace{-10pt}

\begin{tabular}{lccccc}
% \toprule
% \multicolumn{5}{c}{\textbf{Success Rate (\%)} (with MPC Controller) ($\uparrow$)}  \\ 
\toprule
         & Forests & Garage & Indoor & Campus \\ 
\midrule
\textbf{iPlanner} &70.59 $(\frac{12}{17})$ &\textbf{73.91} $(\frac{17}{23})$ &87.10 $(\frac{27}{31})$ & 79.31 $(\frac{23}{29})$\\ 
\addlinespace[0.5em]
\textbf{iKap}     &\textbf{82.35} $(\frac{14}{17})$ &69.57 $(\frac{16}{23})$ &\textbf{93.54} $(\frac{29}{31})$ & \textbf{86.20} $(\frac{25}{29})$\\ 
\bottomrule
\end{tabular}
% \addlinespace[0.5em]
\label{tab:success_rate}
\end{table}

\subsubsection{Real-time Demo on Real Robot}

\begin{figure}[t]
\centering
\includegraphics[width=\linewidth]{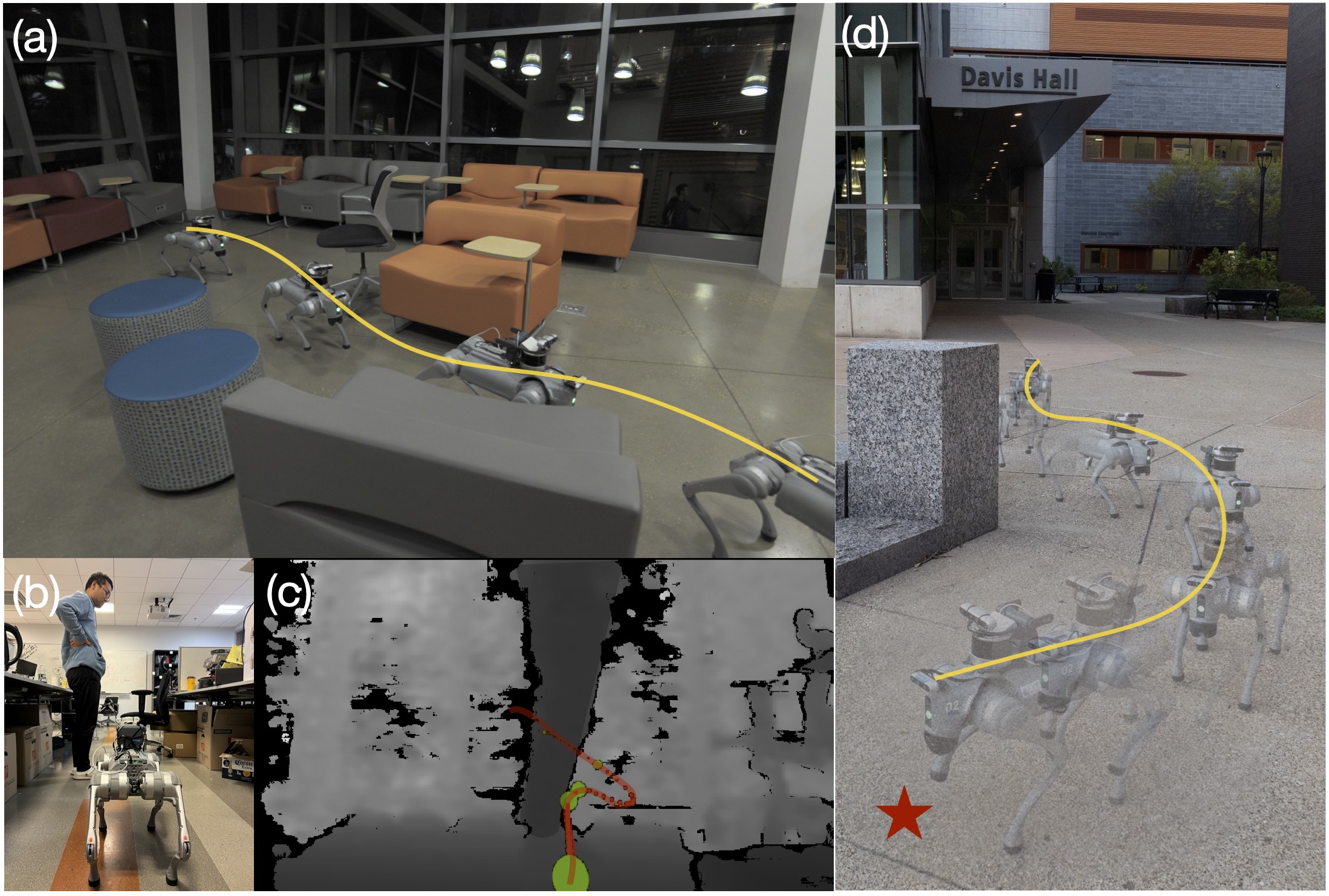}
\caption{
% Snapshots of testing iKap on a legged robot in real-world scenarios including narrowed rooms, open areas with dynamic obstacles like human and static ones like chairs and sofas, and outdoor. (a) The robot dog is commanded with a target goal point 5 meters ahead in the open area with an irregular and partially blocked path; (b) The robot is in between a narrow path with a chair and a human ahead, while (c) shows its depth measurement and the planned path; and (d) The robot is commanded to go from off-road to on-road (red star) with a stone blocking the way.
Snapshots of testing iKaP in real-world scenarios, including narrow rooms, open areas with obstacles, like humans, chairs and sofas, and outdoor environments. (a) The robot is commanded to a goal 5 meters ahead in an irregular, partially blocked open area; (b) The robot navigates a narrow path between a chair and a human; (c) shows its depth measurements and planned path; (d) The robot is commanded to move from off-road to on-road (red star) with a stone blocking the way.
}
\label{fig:realexp}
\end{figure}

To evaluate the real-time online planning capabilities of iKap in real-world environments, we conduct tests using the quadruped robot Unitree GO2 equipped with an Intel RealSense D435i camera for depth perception. The robot operates with bicycle kinematics model, featuring a minimum turning radius of 0.73 meters. The depth measurements are inherently noisy and can suffer from distribution shifts compared to simulation data. Despite these challenges, our planning network, although trained exclusively in simulation without fine-tuning on real-world data, demonstrates strong generalization to real robot deployment. 
As shown in \fref{fig:realexp}, the robot can conduct challenging navigation tasks where the robot starts indoors, passes through a narrow corridor, circles around a public area, and returns to the room along the same path. Additionally, we command the robot to go from off-road to on-road with a large stone blocking it.
% Upon re-entering the room, the robot scans the surroundings before returning to its starting point. 
During this process, the human operator is responsible for sending goal points, while local navigation is autonomously managed by the planner.

To further increase the difficulty of the experiment, we place static obstacles, such as tables and chairs, in the open area of the public space, and have volunteers randomly walk along the robot's path. The experimental results show that when the quadruped encounters obstacles that clearly block its path, it plans a smooth, kinematically feasible trajectory for easier tracking, rather than turning in place to adjust its heading before moving in a straight line. This navigation behavior improves the overall planning speed and enhances the robustness of the trajectory tracking performance by different types of low-level controllers such as PID and MPC.

\section{Conclusion and Limitation} \label{sec:conclusion}
We present iKap, a novel planning system that integrates kinematics into an end-to-end learning pipeline for path planning and control. By combining depth perception with a differentiable MPC module, iKap predicts waypoints while ensuring that the planned trajectories comply with the robot's kinematic constraints.
Experimental results demonstrate that iKap enhances success rates in complex simulated environments and operates effectively under real-world conditions even with noisy depth perception, making it a promising solution for practical robot navigation.
Future work includes evaluating this work across diverse environments beyond human-made structured environments and diverse robot platforms such as Clearpath Husky Observer, as well as extending iKap into three-dimensional space.
% A limitation of this work is that we have only tested our model in human-made structured environments. Exploring iKap in three-dimensional space would also be a valuable direction. 
% In the future, we plan to enhance iKap by developing a time-varying cost map for dynamic obstacles and evaluating its real-world performance across diverse environments.

\section*{Acknowledgements}
This work was partially supported by the DARPA award HR00112490426 and ONR award N00014-24-1-2003. Any opinions, findings, conclusions, or recommendations expressed in this paper are those of the authors and do not necessarily reflect the views of DARPA or ONR.

\bibliography{ref.bib}
\bibliographystyle{IEEEtran}

\end{document}